\documentclass[10pt,twocolumn]{article}

\usepackage{cvpr}
\usepackage{times}
\usepackage{epsfig}
\usepackage{graphicx}
\usepackage{amsmath}
\usepackage{amssymb}
\usepackage{amsthm} 
\usepackage{subfig}
\usepackage{url}
\usepackage{algorithm} 
\usepackage{algorithmic} 

\usepackage{blkarray}

\cvprfinalcopy 

\def\cvprPaperID{65} 

\theoremstyle{plain}
\newtheorem*{ESC}{Epipolar Scales Computation (ESC) Problem}

\newtheorem{proposition}{Proposition}

\providecommand{\norm}[1]{\lVert#1\rVert}

\DeclareMathOperator{\diag} {diag}

\DeclareMathOperator{\rank} {rank}

\DeclareMathOperator{\nullity}   {nullity}

\newcommand{\tr}    	{{\top}}

\renewcommand{\vec}[1]  {\mathbf{#1}}

\newcommand{\al}         {\boldsymbol{\alpha}}

\newcommand{\gradi}     {\ifmmode^\circ\else$^\circ$\fi}

\newcommand{\vc} 	{\vec{c}}

\newcommand{\vzero} 	{\vec{0}}

\begin{document}

\title{On Computing the Translations Norm in the Epipolar Graph}

\author{Federica Arrigoni, Andrea Fusiello\\
\normalsize DIEGM - University of Udine\\ 
\normalsize Via Delle Scienze, 208 - Udine (Italy)\\
\small arrigoni.federica@spes.uniud.it, 
\small andrea.fusiello@uniud.it
\and
Beatrice Rossi\\
\normalsize AST Lab - STMicroelectronics\\
\normalsize Via Olivetti, 2 - Agrate Brianza (Italy)\\
\small beatrice.rossi@st.com
}

\maketitle

\begin{abstract}
This paper deals with the problem of recovering the unknown norm of relative
translations between cameras based on the knowledge of relative rotations and translation directions.
We provide theoretical conditions for the  solvability of such a problem, and we propose a two-stage method to solve it. First, a cycle basis for the epipolar graph is computed, then all the scaling
factors are recovered simultaneously by solving a homogeneous linear system.
We demonstrate the accuracy of our solution by means of
synthetic and real experiments.

\end{abstract}


\section{Introduction}


In this paper we deal with the Epipolar Scales Computation (ESC) problem, namely 
the 
problem of recovering (up to a global scale factor) the
unknown norms -- also referred to as \emph{epipolar scales} --  of the relative
translation directions extracted from the essential matrices.
In fact, only the translation directions can be computed from the epipolar geometries, but not their norms, owing to the well-known depth-speed ambiguity.
The problem can be usefully modeled by considering the  
\emph{epipolar graph}, where nodes are the images and edges correspond to epipolar relationships between them.  

The only solution in the literature \cite{ZelFau96} considers graphs with a special structure. This paper presents a more general technique, together with a formal analysis of the conditions under which the ESC problem is solvable.

The ESC problem finds application in Structure-from-motion (SfM), namely the problem of recovering 3D structure (of the scene) and motion  (of the cameras) from point correspondences.
A paradigm which is gaining increasing attention in the community consists in first computing the \emph{relative} motion of all the cameras and then deriving their
\emph{absolute} position and angular attitude by considering the whole  epipolar graph at the same time.

Almost all these global techniques \cite{Govindu01,MartinecP07,Arie12,Moulon13,ozyesil2013stable} first solve for rotations and then for translations. The problem of the unknown norms is bypassed either by exploiting  implicit or explicit
point triangulation (e.g.~\cite{Arie12, Snavely14, disco, Kahl08, MartinecP07, SinhaSS10}), or
by solving a bearing-only network localization (e.g.~\cite{Govindu01, Brand04, Jiang13, Moulon13, ozyesil2013stable}), where the relative 
translation directions  expressed in an absolute frame are regarded as bearing measures that  globally constraint the position of the cameras. 
Conditions under which positions are recoverable are studied in \cite{ozyesil2013stable} and they refer to the concept of \emph{parallel rigidity} \cite{Whi97}. The ESC problem is very related to this one, although there are some differences: in the ESC problem the input are \emph{relative} rotations and relative translation directions, and the output are relative distances among cameras; in the bearing-only network localization problem the input are \emph{absolute} rotations and relative translation directions, and the output are \emph{absolute} positions of the cameras.

A different approach  to global SfM consists in recovering rotations and translations \emph{simultaneously}, by working on the manifold of rigid motions SE(3). 
This approach, although being more principled, is less explored, probably due to the lack of a general solution to the ESC problem. 
Indeed, essential matrices do not fully specify elements
of SE(3), due to the scale ambiguity in the relative translations. The only approach 
of this type present in the literature is the iterative solution in \cite{Govindu04}, where 
the ESC problem is overlooked, though,  since at each iteration the current estimates of the absolute motions are used to fix the scales of the corresponding relative translations.

In this paper we provide theoretical conditions that guarantee solvability of the ESC problem, and we propose a two-stage method to solve it.
First, a cycle basis for the epipolar graph is computed, then all the scaling factors are recovered simultaneously by solving a homogeneous linear system.
The key observation is that the compatibility constraints associated to cycles can be seen as equations in the unknown scales.
Thus the ESC problem is cast to the resolution of a single linear system, and solvability depends on the algebraic properties of the coefficient matrix.

We consider two variants of our method, which differ for the algorithm used to obtain a cycle basis, namely computing a Fundamental Cycle Basis (FCB) or a Minimum Cycle Basis (MCB). 
Experiments on synthetic and real data show that they both recover the epipolar scales accurately, and the lowest errors are obtained when using a MCB. Moreover, a MCB can be made resilient to outliers, which pays back for its higher computational cost.

The paper is organized as follows.
First, we define the ESC Problem (Section \ref{sec:pb}) and we introduce the background necessary to address it (Section \ref{background}).
Theoretical results about unique solvability of the problem are discussed in Section \ref{theory}, and the derived method is detailed in Section \ref{algorithm}.
Finally, we evaluate the performances of our contributions to the ESC Problem via experiments on synthetic and real data (Section \ref{experiments}).


\section{Problem Definition} \label{sec:pb}

Consider $n$ pinhole cameras that capture the same (stationary) 3D scene.  
Let $M_{ij}$ denote the relative transformation between cameras
$i$ and $j$, which can be represented as an element of the Special Euclidean Group SE(3),
namely the semi-direct product of the Special Orthogonal Group SO(3) with $\mathbb{R}^3$.  
Accordingly, each relative transformation can be expressed as
\begin{equation}
M_{ij} = \begin{pmatrix}
R_{ij} & \mathbf{t}_{ij} \\
0 & 1
\end{pmatrix}
\end{equation}
where $R_{ij} \in SO(3)$ and $\mathbf{t}_{ij} \in \mathbb{R}^3$ respectively
denote the relative rotation and translation between coordinate frames indexed
by $i$ and $j$.

Suppose that only some $M_{ij}$ are known, represented by index
pairs $(i,j)$ in a set $\mathcal{E} \subseteq \{ 1, 2,\dots, n \} \times \{ 1,
2, \dots, n \}$.
Let $\mathcal{G} = (\mathcal{V},
\mathcal{E})$ denote the \emph{epipolar graph} (also known as the \emph{viewing graph} \cite{Levi03}), which has a vertex for each
camera and edges in correspondence of the available pairwise transformations.
$\mathcal{G}$ is a directed finite simple graph with a labeling of its edge set
by elements of $SE(3)$
\begin{equation}
\Lambda : \mathcal{E} \rightarrow SE(3), \quad
\Lambda(i,j) = M_{ij}
\end{equation}
such that if $(i,j) \in \mathcal{E}$ then
$(j,i) \in \mathcal{E}$, and $\Lambda(j,i) {=} \Lambda(i,j)^{-1}$. Hence, 
$\mathcal{G}$ may also be considered as an undirected graph. 
Let $m$ denote the cardinality of $\mathcal{E}$, i.e. the number of edges of the underlying undirected graph.

In practice, the relative transformations $M_{ij}$ are obtained by factorizing
the essential matrices, which are computed from a collection of point matches
across the input images.
Each essential matrix is known up to scale due to the depth-speed ambiguity.
Therefore, there is a scale ambiguity in
the relative translations, i.e. what can be extracted are the relative
translation \emph{directions} $ \mathbf{\hat{t}}_{ij} = \mathbf{t}_{ij} /{\norm{ \mathbf{t}_{ij} }}. $ 
In other words, the scale factors $\alpha_{ij} = \| \mathbf{t}_{ij} \|$ of the relative
translations are unknown.  
Note that the number of such unknowns is $m$ since $\| \mathbf{t}_{ij} \| = \| \mathbf{t}_{ji} \|$.

The goal here is to reduce all the unknown scaling factors of the relative
translations into a \emph{single} global scaling factor, which 
cannot be eliminated.    
In other words, the present work addresses the following problem.
\begin{ESC}
Given the relative rotations $R_{ij} \in SO(3)$ and relative translation directions
$ \mathbf{\hat{t}}_{ij}  \in \mathbb{R}^3$ for $(i,j) \in \mathcal{E}$, compute the scaling factors
$\alpha_{ij} = \norm{  \mathbf{t}_{ij}  }$ of the relative translations up to a single global
scaling factor\footnote{Please note that when referring to a ``unique'' solution  to the ESC problem we will include the global scale indeterminacy.}.
\end{ESC}

In particular, the questions are: under which assumptions the ESC Problem admits solution, and which algorithm
can solve it.
A pair $( \mathcal{G}, \Lambda )$ for which it is possible to solve the ESC problem is called a \emph{solvable epipolar graph}.


\section{Background}
\label{background}

In this section we review some useful concepts from graph theory \cite{bases},
and we describe the Zeller-Faugeras method \cite{ZelFau96}, of which our method is
a generalization.

\subsection{Cycle Bases} 
\label{graph_theory}
 
Consider a finite simple graph $\mathcal{G} =
(\mathcal{V},\mathcal{E})$, where $\mathcal{V}$ is the set of \emph{vertices} (or \emph{nodes}) of cardinality $n$ and
$\mathcal{E}$ is the set of \emph{edges} of cardinality $m$. 
 If the edges are ordered pairs of vertices then $\mathcal{G}$
is a \emph{directed graph}, otherwise $\mathcal{G}$ is an
 \emph{undirected graph}. A \emph{weighted graph} is a graph together with a
 weight function $w : \mathcal{E} \to \mathbb{R}^+$. 
 

$\mathcal{G}$ is called \emph{connected} if there exists a path from each vertex to any other.
$\mathcal{G}$ is called \emph{biconnected} if it has no articulation points,
where a vertex $v \in \mathcal{V}$ is an \emph{articulation point} if $\mathcal{G} \setminus\{v\}$ is disconnected. 
A graph is a \emph{tree} if it is connected and it has $n-1$ edges. A subgraph
of a connected graph $\mathcal{G}$ is called a \emph{spanning tree} if it has the same vertices of $\mathcal{G}$ and it is a tree. 
A single spanning tree of a graph can be found in linear time $O(m+n)$ by either depth-first search or breadth-first search.

A \emph{cycle} in an undirected graph is a subgraph in which every
vertex has even degree, where the \emph{degree} of a vertex is the number of times that the vertex
occurs as the endpoint of an edge.
A cycle is a \emph{circuit} if it is connected and every one of its vertices has
degree two. In this paper we use the notation $(i_1, i_2, \dots, i_{N-1}, i_N)$ to denote the
$N$-length circuit formed by the edges $\{ (i_1,i_2), (i_2,i_3), \dots,
(i_{N-1},i_N), (i_N,i_1) \}$.  



\begin{figure}[!ht]
\centering\includegraphics[width=0.8\columnwidth]{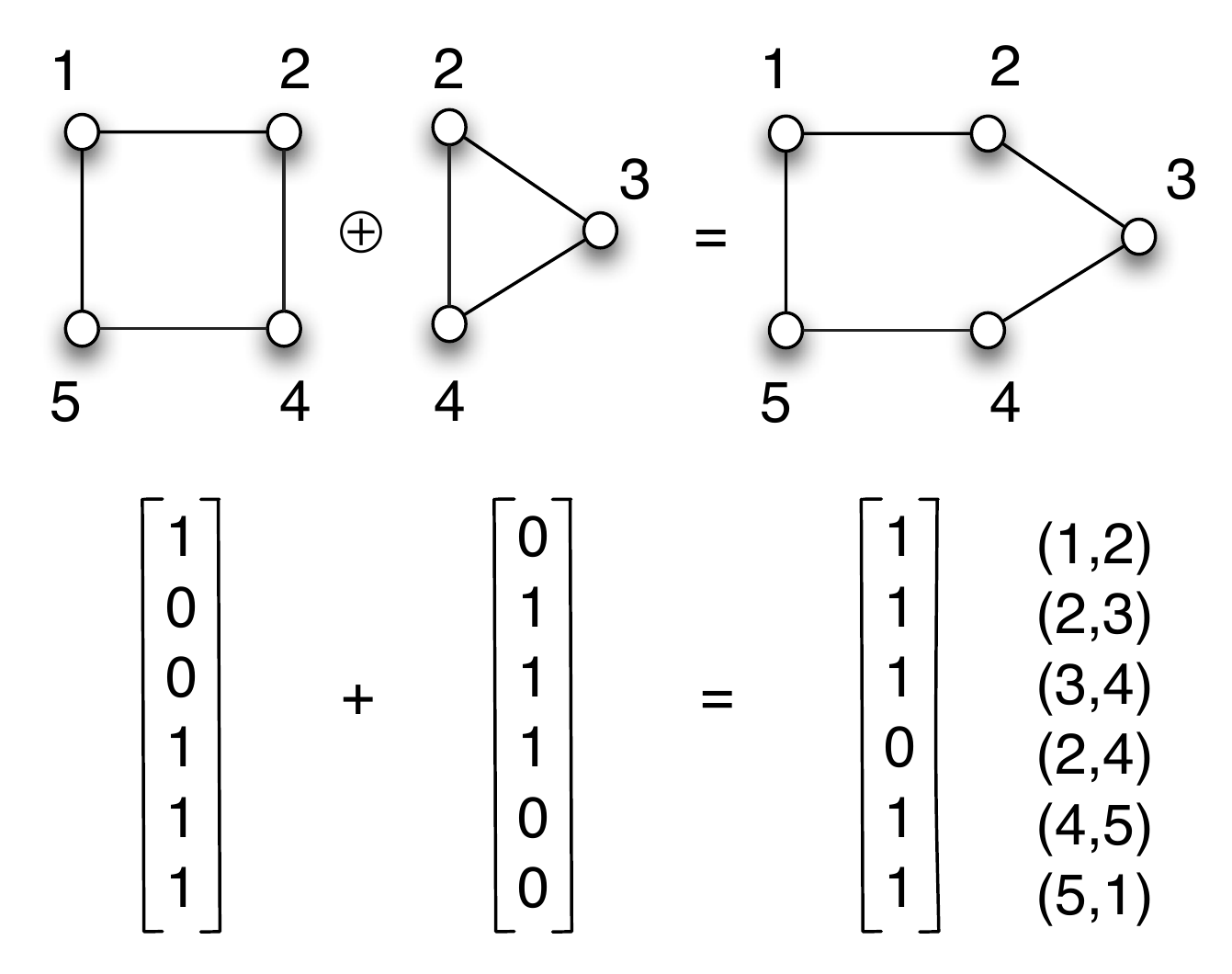}
\caption{The sum of two cycles is a cycle where the common edges vanish. }
\label{figure:sum}
\end{figure}

If $C_1, \dots, C_k$ are cycles of $\mathcal{G}$, then the \emph{sum of cycles}
$C_1 \oplus \dots \oplus C_k$ is defined as the cycle consisting of all the edges that are
contained in an odd number in the cycles $C_i$, as illustrated in Figure \ref{figure:sum}.  A \emph{cycle basis} is a minimal set
of circuits such that any cycle can be written as linear combination of the circuits in the
basis. Viewing cycles as vectors indexed by edges, addition of cycles corresponds to modulo-2 sum of vectors, and the cycles of a graph form a vector space in $GF(2)^{m}$. The dimension of such a space is $m - n + cc$, where $cc$ denotes the number of connected components in $\mathcal{G}$.

If $\mathcal{G}$ is connected and $\mathcal{T}$ is any arbitrary spanning tree
of $\mathcal{G}$, then adding any edge from $\mathcal{G} \setminus \mathcal{T} $  to $\mathcal{T}$ will generate a
circuit. The set of such circuits forms a cycle basis, which is referred to as
\emph{fundamental cycle basis} (FCB). This simple technique for extracting a cycle basis is summarized in Algorithm \ref{alg_spanningtree} and it runs in $O(m+n)$ time.

\begin{algorithm}[h!]
 \caption{Spanning tree}
 \begin{algorithmic}
 \REQUIRE Connected graph $\mathcal{G}=(\mathcal{V},\mathcal{E})$
 \ENSURE Fundamental Cycle Basis $\mathcal{B}$
\begin{enumerate}
\item Initialize $\mathcal{B} = \emptyset$. 
\item Compute a spanning tree $\mathcal{T}$.

\FOR{$(x,y) \in \mathcal{E} \setminus \mathcal{T}$}
\STATE Create the cycle $C(x,y) = P(x,y) \cup (x,y)$, where $P(x,y)$ is the shortest path in $\mathcal{T}$ between $x$ and $y$. Add $C(x,y)$ to $\mathcal{B}$. 

\ENDFOR

\end{enumerate}
\end{algorithmic}
\label{alg_spanningtree}
\end{algorithm}

\begin{figure}[!ht]
\centering

\subfloat[][Epipolar graph.]
{
\includegraphics[width=0.32\columnwidth]{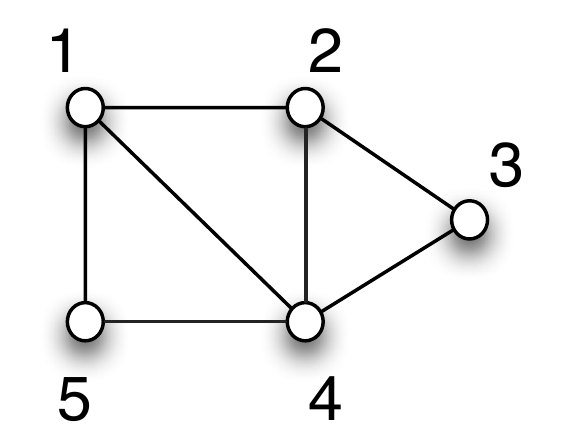} 
} 
\quad \quad
\subfloat[][Minimum cycle basis.]
{
\includegraphics[width=0.5\columnwidth]{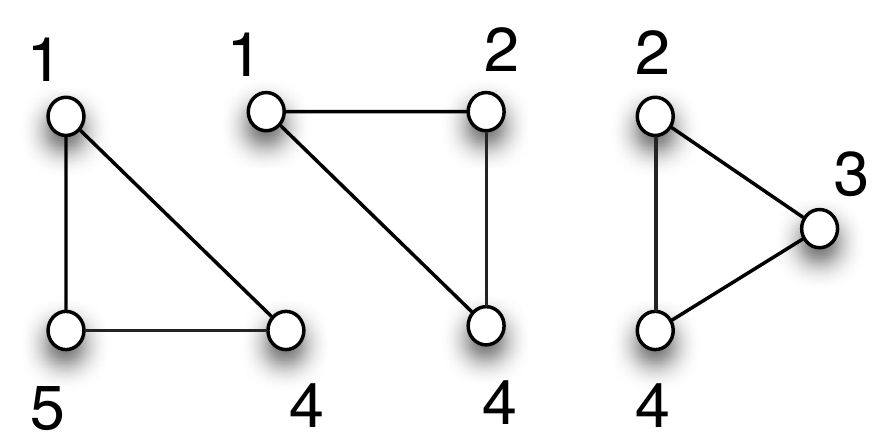} 
} 

\subfloat[][Fundamental cycle basis associated to the spanning tree $\mathcal{T} = \{(1,2), (2,3), (3,4), (4,5) \}$.]
{
\includegraphics[width=0.8\columnwidth]{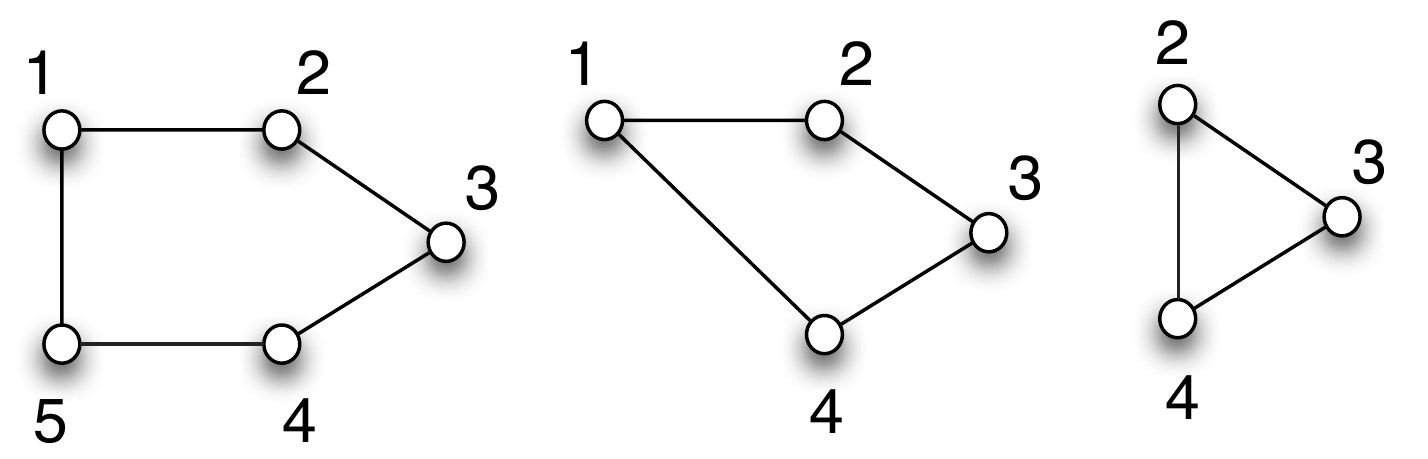} 
} 
\caption{Example of a MCB and a FCB for a given epipolar graph. In general, the latter is composed of longer cycles.}
\label{figure:diff}
\end{figure}

The \emph{length} of a cycle is either the number of edges in the cycle (in
unweighted graphs) or the sum of the weights of the edges in the cycle (in
weighted graphs). A \emph{minimum cycle basis} (MCB) is a basis of total minimum
length. In general a MCB is not unique.  Horton's algorithm \cite{horton} finds a MCB in polynomial time, requiring at most $O(m^3n)$ steps, assuming that the underlying graph is biconnected.
This method is described in Algorithm \ref{alg_horton}.
The last step in Algorithm \ref{alg_horton} can be implemented by applying Gaussian elimination to a $0,1$-matrix whose rows are the vectors in $GF(2)^m$ corresponding to the cycles generated in Step 2.
Figure \ref{figure:diff} outlines the difference between MCB and FCB. 

\begin{algorithm}[h!]
 \caption{Horton}
 \begin{algorithmic}
 \REQUIRE Biconnected graph $\mathcal{G}=(\mathcal{V},\mathcal{E})$
 \ENSURE Minimum Cycle Basis $\mathcal{B}$
\begin{enumerate}
\item Find the shortest path $P(x,y)$ between each pair of 
vertices $x,y \in \mathcal{V}$.
\FOR{$v \in \mathcal{V}$}
\FOR{$(x,y) \in \mathcal{E}$}
\STATE Create the cycle $C(v,x,y) = P(v,x) \cup P(v,y) \cup (x,y)$ and
calculate its length. 
Degenerate cases in which $P(v,x)$ and $P(v,y)$ have vertices other than 
$v$ in common can be omitted. 
 \ENDFOR
\ENDFOR
\item Order the cycles by increasing lengths.
\item Initialize $\mathcal{B} = \emptyset$. Add to $\mathcal{B}$ the 
next shortest cycle if it is independent from the already selected ones.
\end{enumerate}
\end{algorithmic}
\label{alg_horton}
\end{algorithm}

\subsection{Zeller-Faugeras method}  
\label{faugeras}

Our method for solving the ESC Problem is inspired by \cite{ZelFau96}, where
the authors derive the scale factors from the composition of rigid motions.

\begin{figure}[!ht]
\centering\includegraphics[width=0.9\columnwidth]{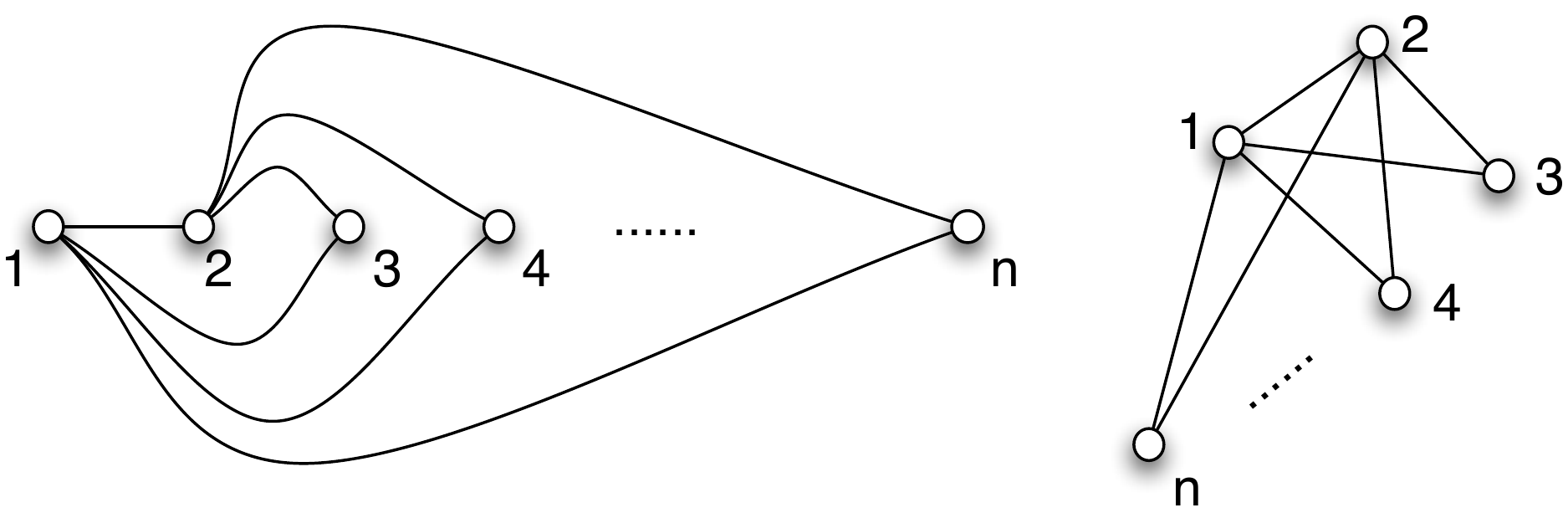}
\caption{The epipolar graph corresponding to the Zeller-Faugeras method
  \cite{ZelFau96}. It is made of $n-2$ circuits of length 3 all sharing a common
  edge.}
\label{figure:sequence}
\end{figure}

If we consider a \emph{sequence} of $n$ images, whose epipolar graph is represented in
Figure \ref{figure:sequence}, then the following compositional rule holds
\begin{equation}
\mathbf{t}_{1 i } = R_{12} \mathbf{t}_{2i} + \mathbf{t}_{12}
\end{equation}
which is equivalent to
\begin{equation}
\alpha_{1i} \mathbf{ \hat{t}}_{1i}  = \alpha_{2i} R_{12} \mathbf{ \hat{t}}_{2i}  +
\alpha_{12} \mathbf{ \hat{t}}_{12} .
\label{eq_faugeras}
\end{equation}
This leads to the following solution for the ratios of the scale factors
\begin{gather} 
 \frac{\alpha_{12}}{ \alpha_{1i}} = \frac{ ( R_{12} \mathbf{\hat{t}}_{2i} \times
   \mathbf{ \hat{t}}_{1i} )^{\mathsf{T}} ( R_{12} \mathbf{ \hat{t}}_{2i}  \times \mathbf{ \hat{t}}_{12}  )
 }{\norm{ R_{12} \mathbf{\hat{t}}_{2i} \times \mathbf{ \hat{t}}_{12} }^2}.
\label{eq:kanatani}
 \end{gather}
More precisely, if we arbitrarily fix the value of (e.g.) $\alpha_{12}$, then we
can compute the remaining scaling factors $\alpha_{1i}$  by using
the equations above. The arbitrary choice of $\alpha_{12}$ corresponds to the
global scaling factor, which can not be computed without external measurements.

Our method can be seen as an extension of this approach to general epipolar
graphs, with a formal analysis of the conditions that guarantee solvability.

\section{Theoretical Results}
\label{theory}

In order to address the ESC Problem, we consider the composition of pairwise motions along circuits, which must return the identity transformation.

We observe that it is impossible to solve the ESC Problem in the presence of edges not belonging to any cycle. 
Indeed, the norm of such edges can be chosen arbitrarily without any impact on the other scales, since they are not constrained by other edges.
For this reason we assume that each edge in $\mathcal{E}$ belongs to (at least) one cycle, namely the set of edges associated to a cycle basis coincides with $\mathcal{E}$ itself. Such a graph is also called \emph{bridgeless}.

\subsection{A single circuit}

For simplicity of exposition, we first consider
the case where the epipolar graph consists of a single circuit $C$ of length $N \ge 3$, e.g. $C=(1,2,\dots,N-1,N)$. 
The composition of the pairwise motions along $C$ yields the $4 \times 4$ identity matrix, namely 
\begin{equation}
 M_{1 2} M_{2 3} \dots M_{{N-1}, N} M_{N1} = I.
\label{compatibility}
\end{equation}
Note that this equation is written by traversing the cycle in a given order (clockwise or anti-clockwise) while considering the \emph{directed} epipolar graph.
Equation \eqref{compatibility} is called the \emph{compatibility constraint}, and it can also be expressed as
$M_{1 2} M_{2 3} \dots M_{{N-1}, N} = M_{1N}$.
Considering separately the rotation and translation terms, it results in
\begin{gather}
 R_{1 2} R_{2 3} \dots R_{{N-1}, N} = R_{1N} \label{compatibility_R} \\
  \alpha_{12} \mathbf{\hat{t}}_{12} + \sum_{k=2}^{N-1} ( \prod_{i=1}^{k-1} R_{i,i+1} )  \alpha_{k,k+1} \mathbf{\hat{t}}_{k,k+1}
  = \alpha_{1N} \mathbf{\hat{t}}_{1N}  
\label{compatibility_t}
\end{gather}
where the relation between translations can be viewed as a homogeneous linear equation in the unknown scales.
Note that the Zeller-Faugeras method considered the
compatibility constraint for $N=3$.

Equation \eqref{compatibility_t} can also be expressed in terms of
differences between the camera centers (i.e. the baselines), if the absolute rotations of the cameras are known.
Let $R_1, \dots , R_n \in SO(3)$ denote the absolute rotations,
let $\mathbf{b}_{ij}  \in \mathbb{R}^3$ denote the baseline joining the optical centers of cameras $i$ and $j$, 
and let $\mathbf{ \hat{b}}_{ij}  \in \mathbb{R}^3$ denote the versor of the baseline $\mathbf{ b}_{ij} $.
Using this additional information, the product of relative rotations in \eqref{compatibility_t} reduces to 
$R_1 R_k^{\mathsf{T}}$. Indeed, the link between relative and absolute rotations in encoded by the formula $R_{ij} = R_i R_j^{\mathsf{T}}$, thus 
all the factors in \eqref{compatibility_t} simplify except of the first and the last one.
By multiplying both sides by $-R_1 ^ {\mathsf{T}}$, we obtain 
\begin{equation}
 - \alpha_{12} R_1^{\mathsf{T}} \mathbf{\hat{t}}_{12} - \sum_{k=2}^{N-1}  \alpha_{k,k+1} R_k^{\mathsf{T}} \mathbf{\hat{t}}_{k,k+1}
  = -\alpha_{1N} R_1^{\mathsf{T}} \mathbf{\hat{t}}_{1N}  
\end{equation}
which coincides with
\begin{equation}
 \sum_{k=1}^{N-1}  \alpha_{k,k+1} \mathbf{\hat{b}}_{k,k+1}
  = \alpha_{1N} \mathbf{\hat{b}}_{1N}  
\label{compatibility_baselines}
\end{equation}
since the baselines are related to the relative translations through the formula $\mathbf{b}_{ij}  = - R_i^T \mathbf{t}_{ij}$.
Note that the baseline versor can be viewed as the direction of the relative translation expressed in the absolute reference frame. For this reason, we can also regard $\mathbf{ \hat{b}}_{ij} $ as the  \emph{bearing} of camera $j$ as seen from camera $i$.

We now discuss under which conditions Equation \eqref{compatibility_t}
gives means to compute the unknown scaling factors $\alpha_{ij}$ (up to a global
scale). 
Let $A \in \mathbb{R}^{3 \times N}$ be the coefficient matrix associated to Equation \eqref{compatibility_t}, whose entries depend on the relative rotations and translation directions, and let
$\al  \in \mathbb{R} ^ N $ be the stack of the scales $\alpha_{ij}$. 
Using this notation, the
compatibility constraint reduces to a homogeneous linear system of the form
$A\al  = 0$.
Thus the ESC Problem admits a unique non-trivial solution -- that corresponds to the 
one-dimensional null space  of $A$ -- if and only if $\rank{(A)} = N - 1$. 
Moreover, we have $\rank{(A)} \le 3$, since $A$ is a $3 \times N$ matrix.
Thus, in a circuit of length $N$ the ESC problem can be solved uniquely  only if $N \le 4$.


Observe that specific motions cause $\rank{(A)}$ to drop.
In particular, $\rank{(A)}=1$ if and only if the camera centers are collinear and $\rank{(A)} = 2$ if and only if the camera centers lie on a common plane.

This implies that for $N=4$ the ESC problem has a unique solution provided that the cameras are in a general position, otherwise multiple solutions are possible.  
On the contrary, for $N=3$, the camera centers \emph{must} be coplanar (as it is the case if we assume correct measurements), otherwise the problem admits only the trivial solution $\al  = 0$. If the centers are collinear then multiple solutions arise.

\subsection{A generic epipolar graph}

We now consider a generic epipolar graph, providing conditions for the ESC Problem to admit a unique solution.
We have just shown that -- if $\mathcal{G}$ is formed by a single circuit -- it is possible to recover the epipolar scales if and only if its length is $3$ or $4$ (provided that the cameras are in a general configuration). Thus a circuit of length $N \ge 5$ is not solvable \emph{alone}, because the associated linear system yields multiple solutions. However, when several cycles are considered in a generic epipolar graph, it might be possible to recover the scaling factors also in the presence of circuits of length $N \ge 5$.

\begin{figure}[!htbp]
\centering
\subfloat[][The epipolar graph contains a circuit of length 5.]
{
\includegraphics[width=0.4\columnwidth]{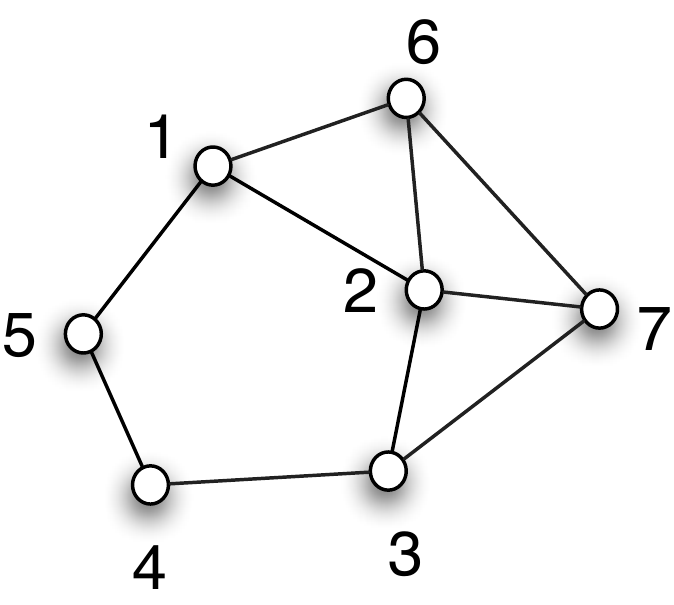} 
\label{fig_5circuit}
}
\quad \quad
\subfloat[][The epipolar graph is not biconnected.]
{
\includegraphics[width=0.35\columnwidth]{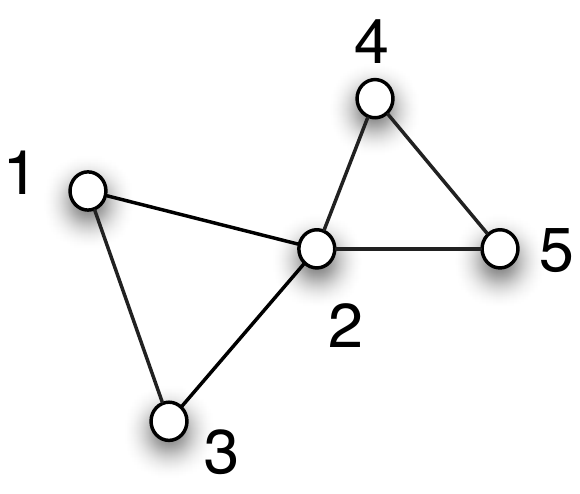} 
\label{fig_no_biconnected}
}

\caption{Examples of a solvable epipolar graph (left) and of an unsolvable epipolar graph (right).}
\end{figure}

To see this, consider the case of Figure \ref{fig_5circuit}.
The key observation is that the $5$-length circuit has two edges in common with a solvable subgraph of $\mathcal{G}$.
Specifically, the epipolar scales can be recovered as follows by considering the circuits $(1,6,2)$, $(2,6,7)$, $(2,7,3)$ and $(1,2,3,4,5)$.
First, we arbitrarily choose the scaling factor of an edge of the circuit $(1,6,2)$, and compute the remaining scales by solving the associated linear system, which has a unique solution since it has length 3.
This cycle shares the edge $(2,6)$ with the $3$-length circuit $(2,6,7)$. We use such an edge to fix the global scaling factor of $(2,6,7)$, and solve for the remaining scales. 
The same happens when considering the $3$-length circuit $(2,7,3)$.
In this way the scales of the edges $(1,2)$ and $(2,3)$ are already determined when considering the $5$-length circuit $(1,2,3,4,5)$, and only $3$ unknowns
remain, which can be recovered as in a  circuit of length $4$.

An example of an unsolvable epipolar graph is reported in Figure \ref{fig_no_biconnected}, where the graph is not
biconnected. The circuits (1,2,3) and (2,4,5) do not have any edge in
common, thus we can solve separately the ESC Problem for each circuit, but two
unknowns remain which can not be reconciled to a single global scaling
factor. 
It is straightforward to see that this generalizes to all the situations where articulation points are
present, as mentioned also \emph{en-passant} in \cite{Moulon13}. In other words, the following proposition holds.

\begin{proposition}\label{prop:nec}
The ESC Problem admits a unique solution  only if the epipolar graph is biconnected.
\end{proposition}

Note that the requirement of being biconnected avoids both the situation of Figure \ref{fig_no_biconnected} and the presence of edges not belonging to any cycle (biconnected $\Rightarrow$ bridgeless).

It is straightforward to see that the necessary condition of
Proposition \ref{prop:nec} is not sufficient. (For instance, a single
$5$-length circuit is biconnected but the associated linear system admits
multiple solutions). However, it gives a simple condition to detect non solvable
graphs. Accordingly, if the epipolar graph is not biconnected, then our analysis applies to the largest biconnected component of $\mathcal{G}$.

We now provide a necessary and sufficient condition for the ESC Problem to admit solution.
Let $r$ be the total number of circuits present in the graph $\mathcal{G}$.
Each circuit gives rise to a homogeneous linear equation of the form
\eqref{compatibility_t}. 
All these equations can be stacked together to form a matrix
$A$ of dimensions $3 r \times m$, whose entries depend on the relative rotations
and translation directions.
Each triplet of rows in $A$ corresponds to a circuit, while each column corresponds to a relative translation.
In this way all the edges are considered (since each edge belongs to at least one cycle by assumption) and all the existing constraints on the scales are taken into account (since we are considering all the circuits).

Thus the ESC Problem is equivalent to the resolution of a single homogeneous linear
system 
\begin{equation}
A \al  = 0 
\label{eq_system}
\end{equation}
where $\al  \in \mathbb{R}^m$ is the stack of the scaling factors $\alpha_{ij}$. 
In other terms, unique solvability depends on the algebraic properties of the coefficient matrix $A$.
More precisely, the ESC problem admits a unique solution if and only if $\nullity{(A)} = 1$, i.e. if and only if $\rank{(A)} = m-1$.
Such a solution is the $1$-dimensional null-space of $A$, and it can be found by computing the
eigenvector with zero eigenvalue of the matrix $A^{\mathsf{T}} A$. 
This discussion is summarized in the following proposition.

\begin{proposition}\label{bigprop}
Let $A \in \mathbb{R}^{3r \times m}$ be the coefficient matrix constructed by stacking the compatibility constraints associated to all the circuits in $\mathcal{G}$, where $r$ is the number of such circuits.
Let $\al  \in \mathbb{R} ^ m $ be the stack of the scales $\alpha_{ij}$.
The ESC problem admits a unique (non-trivial) solution 
if and only if $\rank{(A)} = m-1$.
\end{proposition}







Note that if an articulation point is present -- as in the case of Figure \ref{fig_no_biconnected} -- then the matrix $A$ can be partitioned into two independent blocks having both rank maximum minus $1$. 
Thus the rank of the whole matrix is $m-2$, i.e. the ESC Problem admits multiple solutions, according to Proposition \ref{prop:nec}.

\paragraph{Local vs global frames.}

Equation \eqref{eq_system} can also be written in terms of the baselines (or bearings), generalizing Equation \eqref{compatibility_baselines}. 
In fact, the equation provided by a circuit $C_k$ can be expressed as
\begin{equation} B \diag(\mathbf{c}_k^\tr) \; \al  = \vzero 
\label{eq:genbaseline}
\end{equation}
where $\mathbf{c}_k$ is the $m \times 1$ indicator vector of the circuit $C_k$, and $B$ is a $3 \times m$  matrix  whose columns are the  baseline versors.
Please note that this equation is written by traversing $C_k$ in an  arbitrary cyclic order (clockwise or anti-clockwise), hence the entries of $\mathbf{c}_k$ have a sign that indicates whether the corresponding edge is traversed 
along the direction specified by its versor (the $k$-th column of $B$), or not.

Equivalently, we can use the Khatri-Rao \cite{KatRao68} product
$\odot$ and write
\begin{equation}
(\vc_k^\tr \odot B) \;   \al  = \vzero .
\end{equation}
In this way we can stack the equations coming from 
 $r>1$ circuits, obtaining \label{eq:CkrB}
\begin{equation}
( C \odot B )  \;  \al  = \vzero 
\label{eq_kr}
\end{equation}
where $C$ is the $r \times m$ stack of the rows $\mathbf{c}_k^\tr$.

Please observe that the matrix $( C \odot B )$ is \emph{not} equal to $A$, but it has the same size and the same null-space (in the noise-free case). Each row in $A$ is of the form
\begin{equation}
(\vc_k^\tr \odot - R_k B) \;   \al  =  -  R_k (\vc_k^\tr \odot  B) \; \al   =    \vzero\end{equation}
where $R_k$ is a rotation that takes into account the fact that in each circuit an arbitrary local reference system has been considered. Hence, there exists a choice of rotations $R_1, \ldots, R_m$ such that 
\begin{equation}
\begin{bmatrix}
-R_1 (\vc_1^\tr \odot  B)\\
\vdots\\
-R_m (\vc_m^\tr \odot  B)
\end{bmatrix}
=
\begin{bmatrix}
-R_1 &  &  \\
  & \ddots & \\
  & & -R_m \\
\end{bmatrix} (C \odot B)
=  A.
\end{equation}



In summary, the equations involving the bearings and those involving the relative motions  are equivalent in terms of constraints on the solution, however they configure two different approaches. 
The bearings in \eqref{eq:genbaseline} require to compute the absolute rotations \emph{before} the scale factors, and the problem gets very close to the bearing-only network localization.
On the other hand, the equations in \eqref{eq_system} -- which give our solution to the ESC problem -- are written with respect to  independent local frames, thereby avoiding the need to solve for the absolute rotations beforehand.

Nevertheless, it might be sometime useful to express the constraints in the ``bearing form'', for it simplifies the discussion, as in the following paragraph.

\paragraph{How many circuits?}
As a matter of fact, considering \emph{all} the circuits is 
redundant. The following result states that what is actually required is a set of
\emph{independent} circuits.


\begin{proposition}\label{prop_independence}
Let $C_1$, $C_2$, $C_3$ be three circuits in $\mathcal{G}$ that satisfy $C_1 \oplus C_2=C_3$. 
Then the equation obtained from the circuit $C_3$ is a linear combination of 
the equations obtained from the circuits $C_1$ and $C_2$.
\end{proposition}

\begin{figure}[!ht]
\centering\includegraphics[width=0.8\columnwidth]{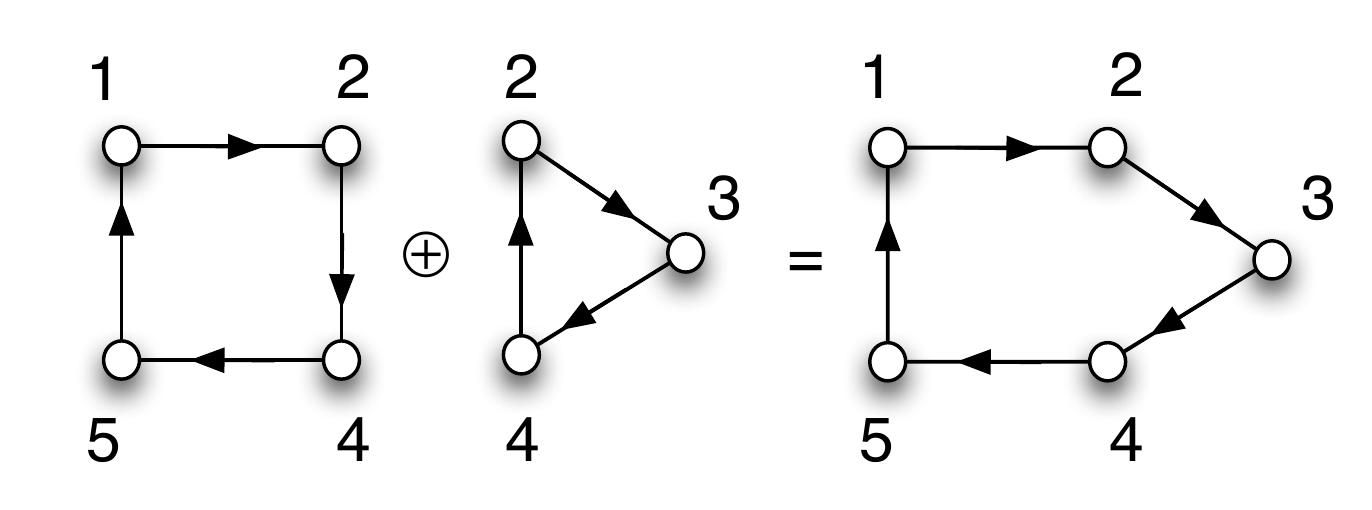}
\caption{Sum of two circuits where the edge in common is traversed in opposite directions.}
\label{figure:sum_order}
\end{figure}

Let us consider two circuits $C_1$ and $C_2$ that share one or more edges,
and let $\vc_1, \vc_2$ be their \emph{signed} indicator vectors. 
The sum of the equations derived from $C_1$ and $C_2$ writes
\begin{equation} B \diag(\vc_1^\tr + \vc_2^\tr)  \al  = 0.
\end{equation}
Without loss of generality let us assume that $C_1$ and $C_2$ are traversed with a cyclic order such that the common edges to $C_1$ and $C_2$ are traversed in opposite directions, as in the case of Figure \ref{figure:sum_order}.
Thanks to this assumptions the entries corresponding to common edges vanishes in  $(\vc_1^\tr + \vc_2^\tr)$, and this is exactly the \emph{signed} indicator vector of $C_1 \oplus C_2$.



Thus, including a
circuit which is the sum of other circuits does not add any independent constraint on the scaling factors.


\section{Proposed Method}
\label{algorithm}

An immediate consequence of Proposition \ref{prop_independence} is that we can consider a \emph{cycle basis} rather than the set of all the circuits in Equation \eqref{eq_system}.
Thus the epipolar scales can be recovered through the following steps.
\begin{enumerate}
\item Compute a cycle basis $\mathcal{B}$ for the epipolar graph by using either Algorithm \ref{alg_spanningtree} or Algorithm \ref{alg_horton}.
\item Construct the $ 3 (m - n + 1) \times m $ coefficient matrix $A$ by stacking the compatibility constraints associated to the cycles in $\mathcal{B}$. 
If $\rank{(A)} = m-1$ then compute the unknown scales by solving system \eqref{eq_system}. Otherwise, it is impossible to find a unique solution to the ESC Problem.
\end{enumerate}
In this way, all the translation norms are recovered simultaneously (up to a global scale) by solving a single homogeneous linear system.
Note that in order to guarantee solvability of the ESC problem, the number of rows in $A$ must be greater than (or equal to) $m-1$, i.e. the following necessary condition must be satisfied
\begin{equation}
m \ge \frac{3}{2} n - 2.
\end{equation}

In the presence of noise unique solvability reduces to test if $A$ has approximately rank $m-1$.
In this case, system \eqref{eq_system} is solved in the least-squares sense, by computing the least eigenvector of the matrix $A^{\mathsf{T}} A$, or -- equivalently -- the least right singular vector in the Singular Value Decomposition (SVD) of $A$. 

Note that system \eqref{eq_system} is \emph{sparse}, since each row contains exactly $N$ non-zero entries, if $N$ is the length of the current circuit.
Thus employing sparse eigen-solvers (such as \textsc{Matlab}  \texttt{eigs}) increases the efficiency of the method.

\paragraph{Which cycle basis?}

In the ideal (noise-free) case any cycle basis returns the desired solution.
Thus the question is which basis is more suitable to our application when relative rotations and translation directions are corrupted by noise and outliers.

Intuitively, the performances of our method with respect to noise are better when using
the shortest circuits, because this limits error accumulation. Therefore a MCB (Algorithm \ref{alg_horton}) should be preferred, because  
a MCB is characterized by the property that no circuit can be the sum of shorter circuits \cite{vismara97}.
This does not hold for a fundamental cycle basis, which in general is composed of longer circuits.

Another advantage of using Algorithm \ref{alg_horton} is that it can easily incorporate robustness to outliers among relative motions. 
Specifically, we take advantage of the redundancy of circuits generated in Step 2, without increasing the computational cost.

We say that a circuit in the epipolar graph is \emph{null} if the composition of the relative rotations along it is equal to the identity. Non null circuits arise when one or more edges are outliers, they provide inconsistent constraints on the epipolar scales, and thus they cannot be part of the cycle basis.
As a consequence, we modify Algorithm \ref{alg_horton} by considering only \emph{null circuits} in Step 2, while the remaining cycles are discarded.

In particular, a circuit $C=(1,2,\dots,N-1,N)$ generated in Step 2 of Algorithm \ref{alg_horton} is kept if the following condition is satisfied
\begin{equation}
 d ( R_{1 2} R_{2 3} \dots R_{{N-1}, N} R_{N1} , I ) \le \epsilon \sqrt{N}
 \label{eq_out_removal}
\end{equation}
where $d(\cdot,\cdot): SO(3) \times SO(3) \mapsto \mathbb{R}^+$ is a bi-invariant metric and $\epsilon$ is a given threshold.
Note that this is an heuristic for finding a cycle basis for a \emph{consistent} subgraph of $\mathcal{G}$, i.e. a subgraph containing only null cycles. 
Thus the number of circuits returned by this version of Algorithm \ref{alg_horton} will be lower than $m-n+1$, in general.


Compared to Algorithm \ref{alg_spanningtree}, Horton's algorithm has a higher computational cost, but this is balanced by increased accuracy and the possibility to discard outliers \emph{while} computing the cycle basis.
In contrast, Algorithm \ref{alg_spanningtree} generates only a minimum set of circuits, thus robustness can be achieved only by rejecting  outliers \emph{before} computing the basis.
Available approaches for detecting outliers include \cite{bayesian,OlssonE11a,EnqvistKO11,Moulon13,isprsarchives-XL-5-63-2014,Snavely14}.
These techniques are computationally demanding and speed is always traded off with accuracy. Moreover, some of them  \cite{bayesian,EnqvistKO11,Moulon13,isprsarchives-XL-5-63-2014} are based anyway on the detection of non-null cycles.

\begin{figure*}[!htbp]
\centering
\includegraphics[width=0.29\linewidth]{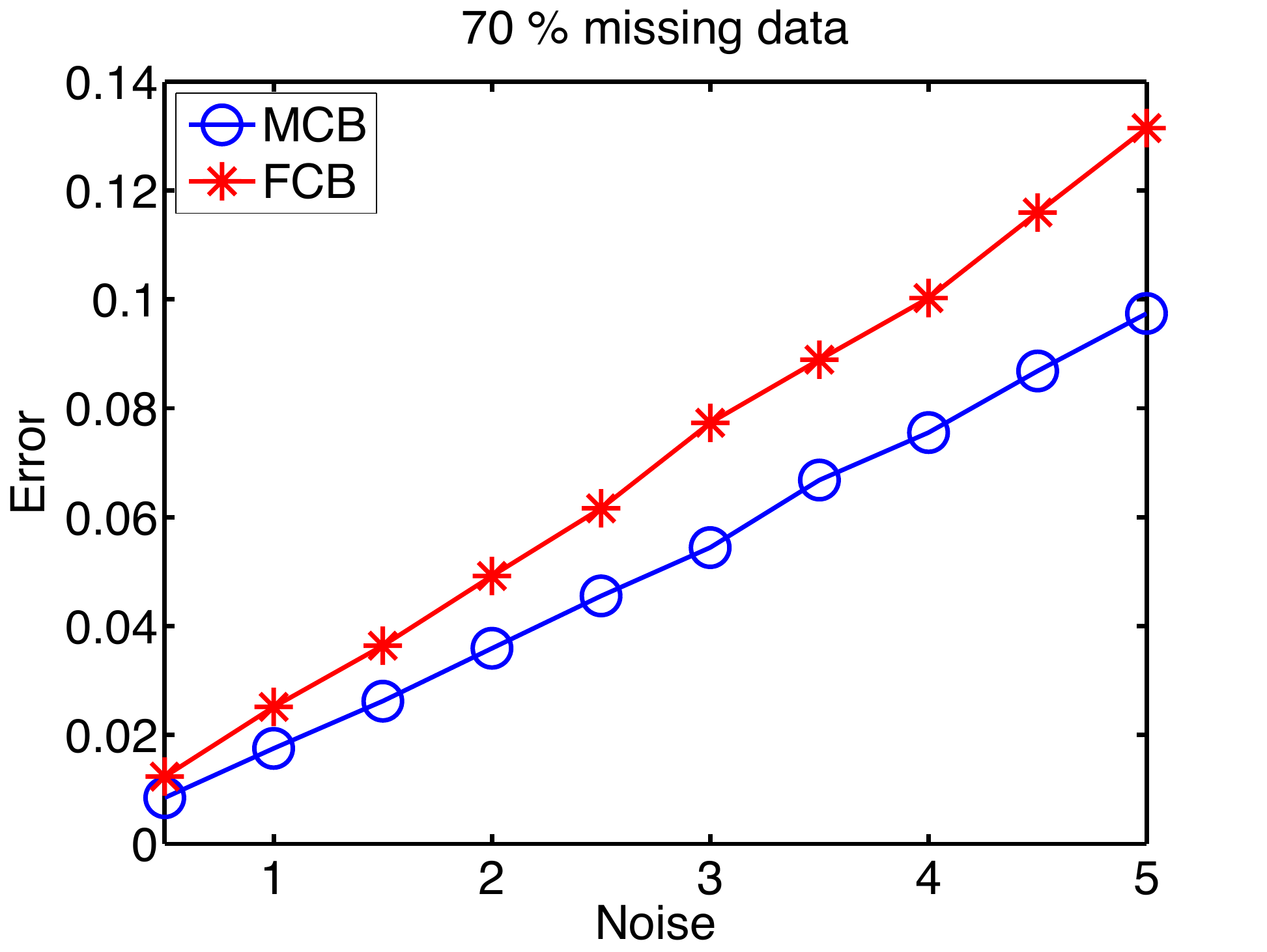} 
\includegraphics[width=0.29\linewidth]{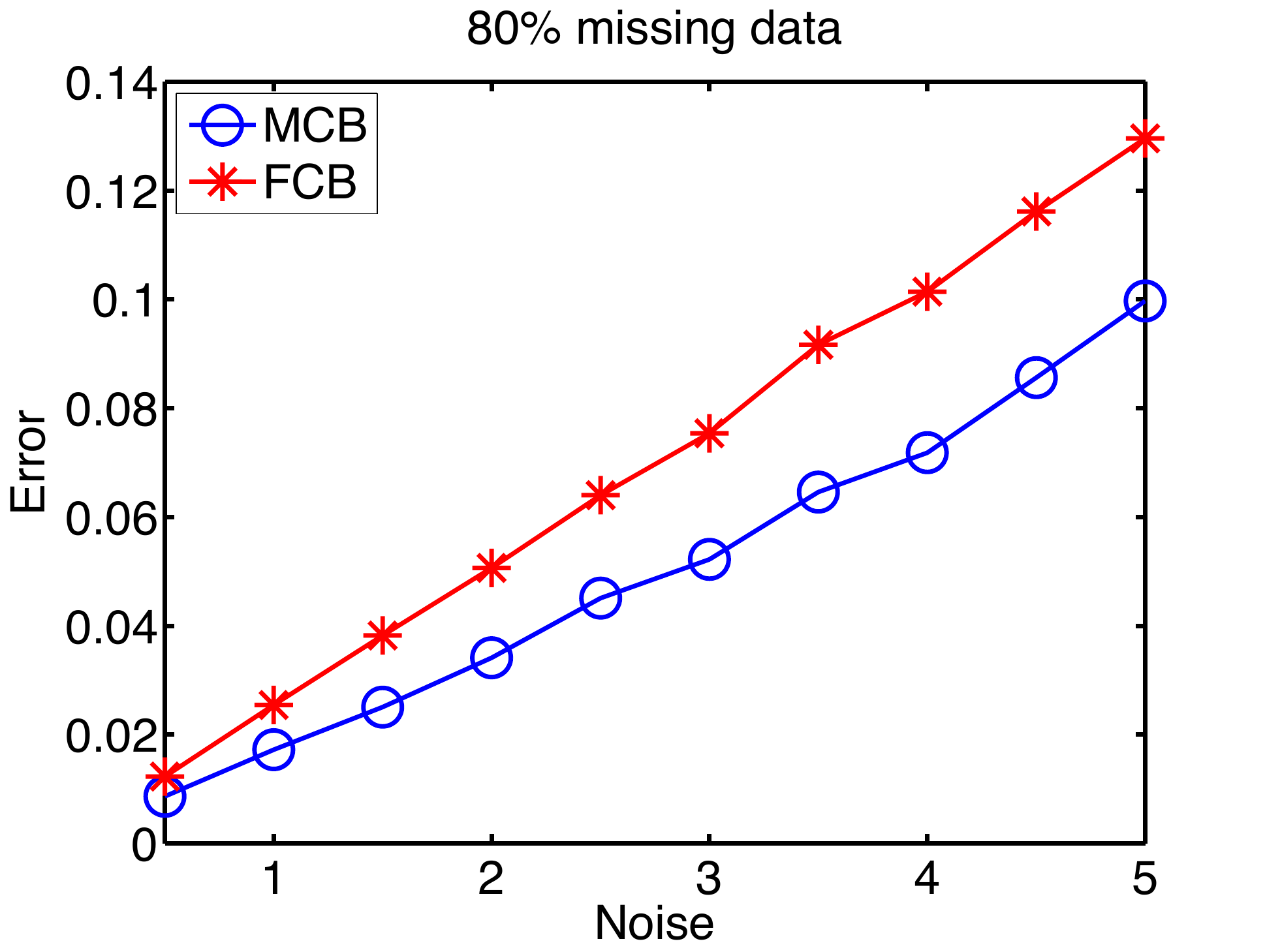} 
\includegraphics[width=0.29\linewidth]{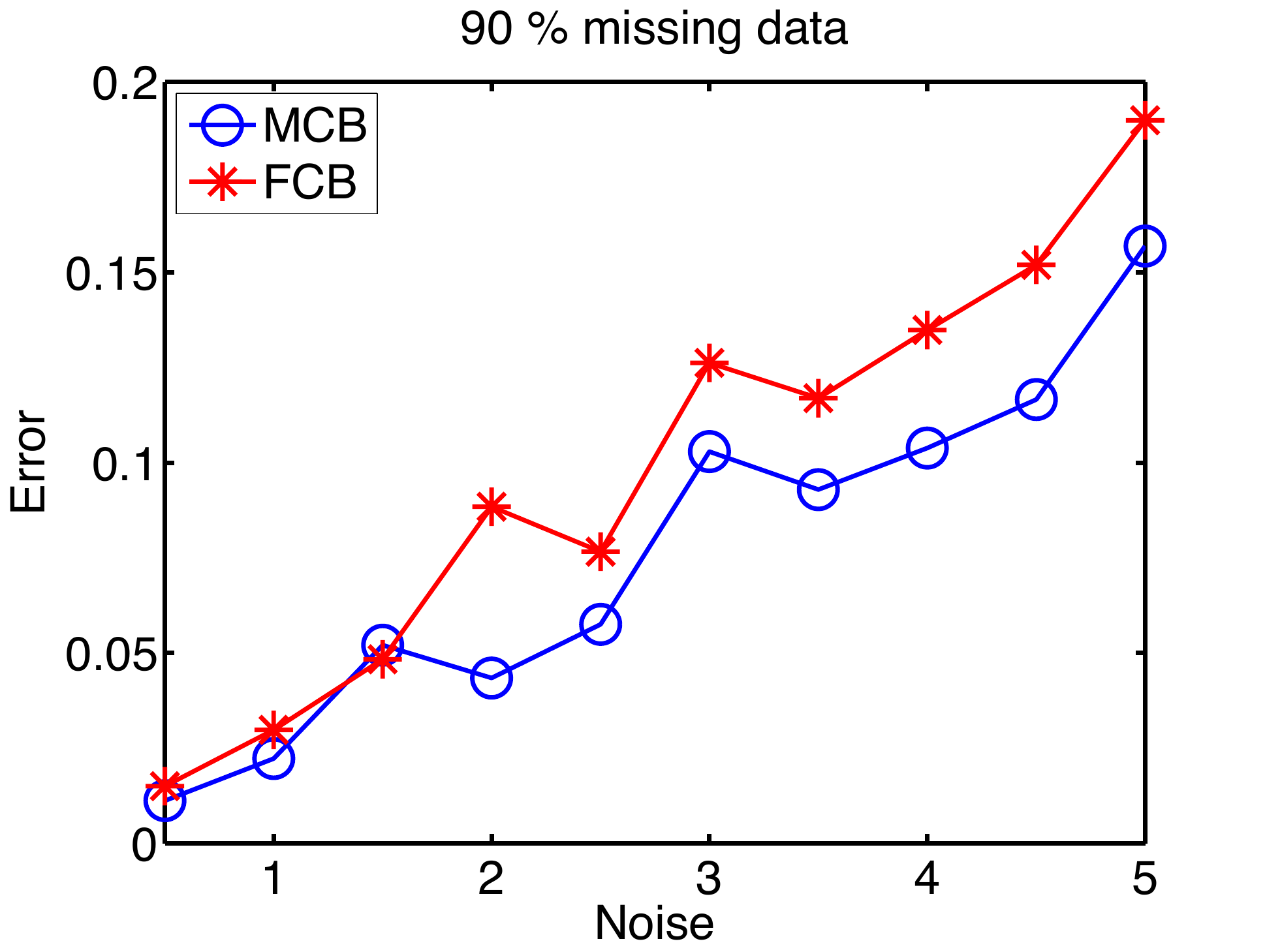} 
\caption{Relative mean error on the scale factors vs standard deviation of noise, for different percentages of missing pairs.}
\label{exp:noise}
\end{figure*}

\begin{figure*}[!htbp]
\centering
\includegraphics[width=0.29\linewidth]{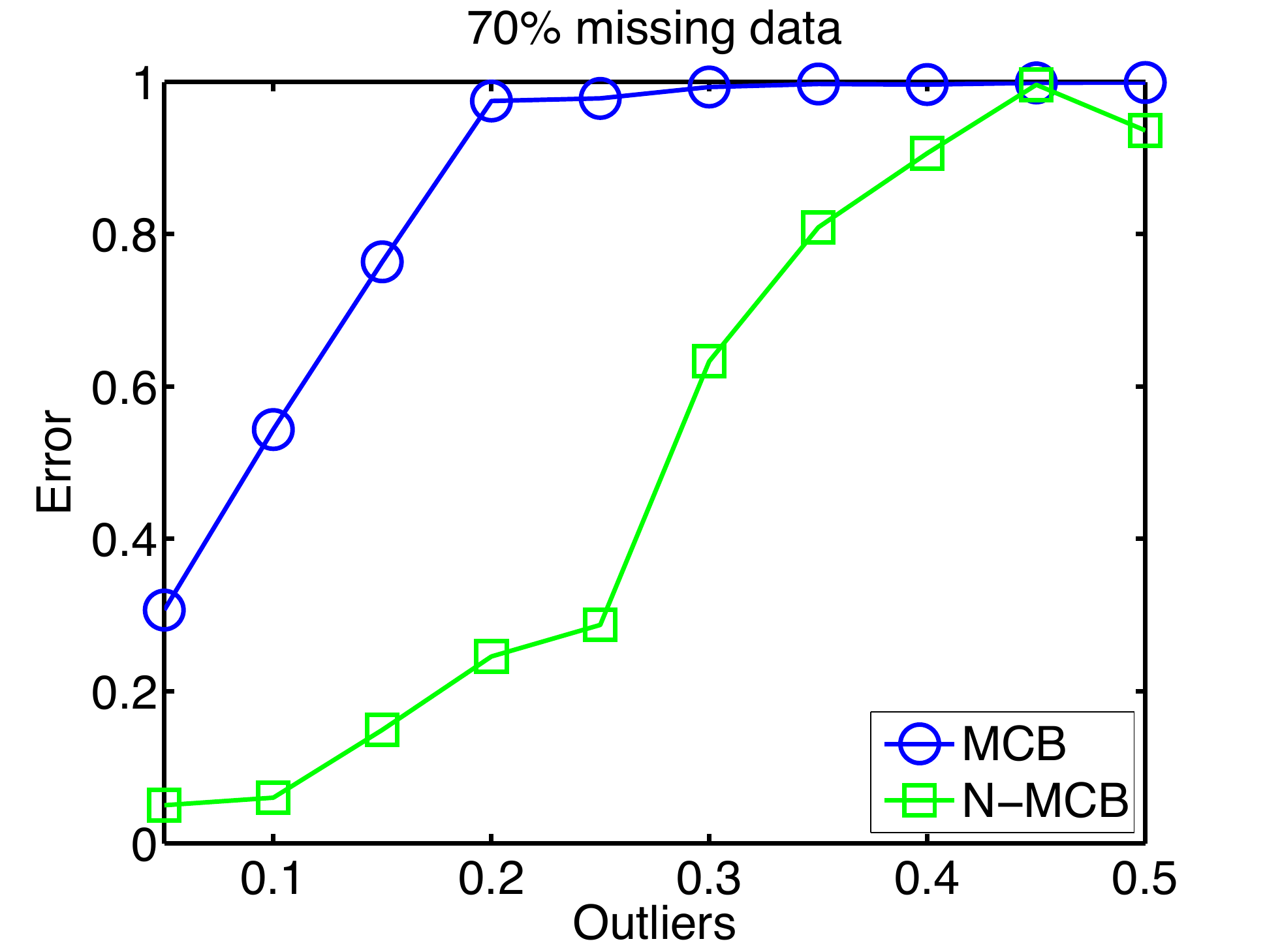} 
\includegraphics[width=0.29\linewidth]{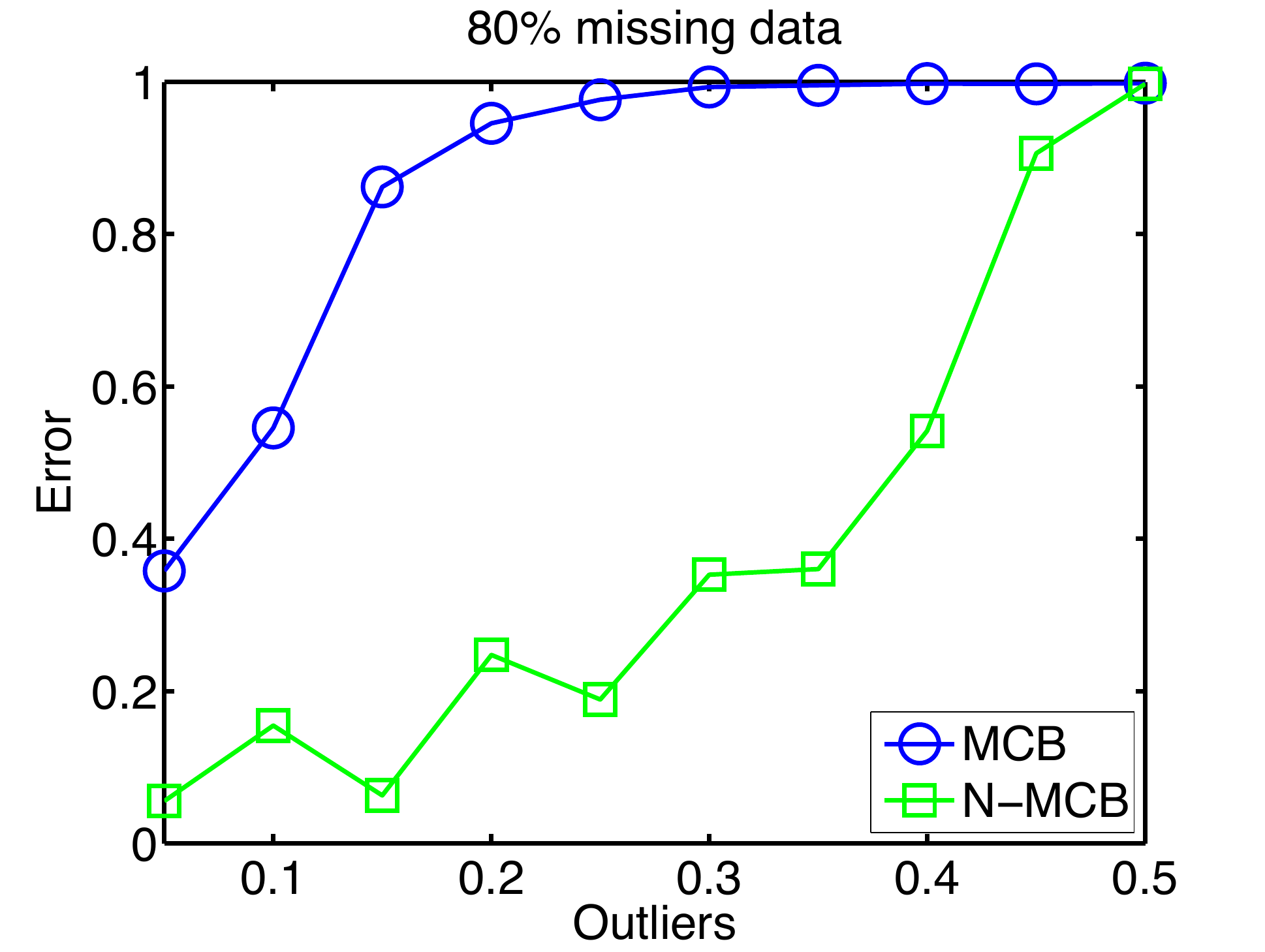} 
\includegraphics[width=0.29\linewidth]{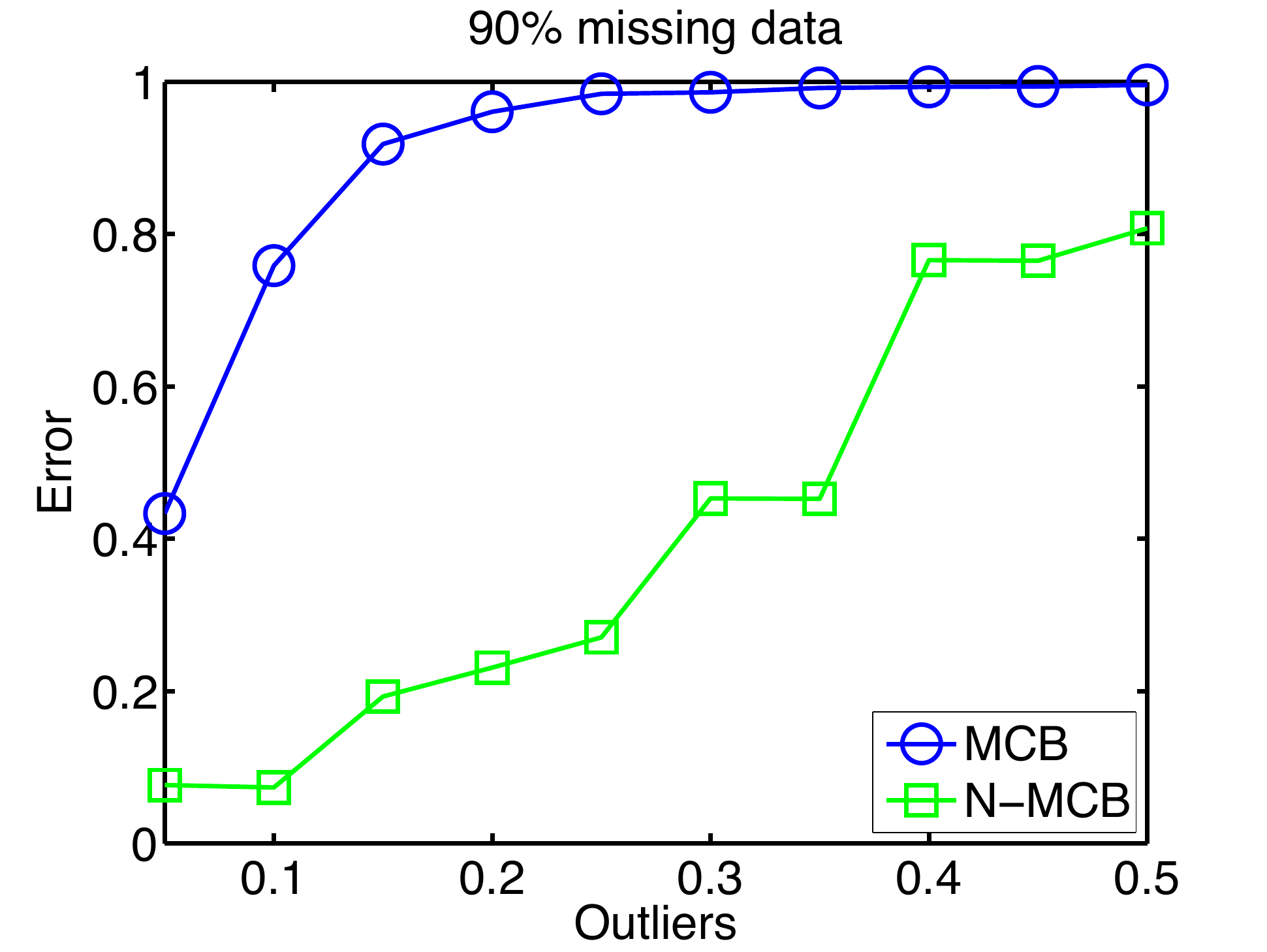} 
\caption{Relative mean error on the scale factors vs fraction of outliers, for different percentages of missing pairs.}
\label{exp:outliers}
\end{figure*}

\section{Experiments}
\label{experiments}

In this section we evaluate our approach on synthetic and real data, analyzing both accuracy and robustness to outliers. 
All the experiments are performed in Matlab on a dual-core 1.3 GHz machine. 
The code is available at \url{www.diegm.uniud.it/fusiello/demo/gmf/}.

\paragraph{Synthetic Data.}
 
We consider $n=100$ cameras where absolute rotations $R_i \in SO(3)$ are sampled from random Euler angles, 
and the $x,y,z$-components of absolute translations $\mathbf{t}_i \in \mathbb{R}^3$ follow a standard Gaussian distribution.
The edge set $\mathcal{E}$ of the epipolar graph is sampled at random.
The available pairwise motions are computed as
$R_{ij} = R_i R_j^{\mathsf{T}}$
and
$\mathbf{t}_{ij} = - R_i R_j^{\mathsf{T}} \mathbf{t}_j + \mathbf{t}_i$.
All the instances simulated correspond to solvable epipolar graphs.
The relative translation directions $\mathbf{t}_{ij}/ \norm{\mathbf{t}_{ij} }$ are corrupted by noise considering their representation in spherical coordinates, so as to remain on the unit sphere.
Specifically, the spherical angles are corrupted by additive Gaussian noise with zero mean and standard deviation $\sigma \in [0.5^{\circ},5^{\circ}]$.
The same perturbation is applied to the relative rotations, considering the angle-axis representation of $SO(3)$. 
All the results are averaged over $10$ trials.

Theoretically, the estimated scales $\tilde{\al}$ should coincide with the
ground truth ones $\al $ up to a multiplicative constant $s \in \mathbb{R}$, namely $\al = s \tilde{\al }$. We estimate such a
constant in the least-squares sense, and we divide the mean of the residuals $r_{ij} = | \alpha_{ij}  - s \tilde{ \alpha }_{ij} | $ by the mean of $\al $, to obtain a \emph{relative mean error} on the scaling factors.

Figure \ref{exp:noise} reports the relative mean errors on the epipolar scales as a function of $\sigma$.
In this experiment we evaluate both Algorithm \ref{alg_spanningtree}, in which a fundamental cycle basis (FCB) is extracted, and Algorithm \ref{alg_horton}, in which a minimum cycle basis (MCB) is computed. The former is highly dependent on the chosen spanning tree, thus for each trial we further average the results over $10$ spanning trees simulated at random.

Both our methods give an accurate
solution to the ESC Problem as noise increases, however the best resilience to noise is achieved by the MCB, as conjectured in the previous section.
In the case of $90\%$ of missing data (right sub-figure) the graph is very sparse and the effect of randomness is amplified, thus producing irregular lines.

We now study the resilience to outliers of our variant of Algorithm \ref{alg_horton} -- henceforth dubbed ``Null MCB'' (N-MCB)  -- in which only null cycles are kept in Step 2.
In this experiment we consider a fixed level of noise ($\sigma = 3^{\circ}$), while
the fraction of wrong relative motions -- randomly generated -- varies from $5\%$ to $50\%$. 
This percentage refers to the available pairwise motions (not to the complete epipolar graph), i.e. the number of outliers is a fraction of $m$.

Figure \ref{exp:outliers} reports the relative mean errors on the epipolar scales as a function of the fraction of outliers,
obtained by MCB and N-MCB. While the former is non robust, the latter shows good resilience to rogue input, confirming the effectiveness of our heuristic for outlier handling.
In particular, the lines corresponding to MCB converge to one since the scale factors converge to zero, thus indicating a complete failure.
The lines of N-MCB are irregular due to the randomness of the data, which is amplified by the presence of both outliers and a high level of missing data.

In this experiment we also analyze the performance of N-MCB in terms of misclassification rate, which is the fraction of effective outliers that are not removed. 
In all the trials we obtain a misclassification rate below $5\%$, thus our heuristic performs well as an outlier detector.


\begin{table}
\centering
\caption{Relative mean errors on the scale factors.
 \label{tab_real}}
 
\scalebox{0.9}{

\begin{tabular}{ l c c c c } 
\hline\noalign{\smallskip}
 & $\%$ missing & FCB &   MCB & N-MCB \\
\noalign{\smallskip}
\hline
\noalign{\smallskip} 
Castle-P30	& 60 &	0.0990 	& 0.0572	& \textbf{0.0326}	 \\
Castle-P19	& 43 &	0.1872	& 0.0707	& \textbf{0.0359}	 \\
Entry-P10		& 18 &	0.0402     & 0.0400	& \textbf{0.0124}	 \\
Fountain-P11	& 2   &	0.0024	& \textbf{0.0017}	& \textbf{0.0017}	 \\ 	
HerzJesu-P25	& 62 &	0.0808	& 0.0312	& \textbf{0.0044}	 \\
HerzJesu-P8	& 18 &	\textbf{0.0040} 	& 0.0108	& 0.0108	 \\
\hline
\noalign{\smallskip}
Average &	& 0.0689 & 0.0353 & \textbf{0.0163} \\
\hline
\end{tabular}
}
\end{table}

\paragraph{Real Data.}

We now consider the EPFL benchmark \cite{real_data}, a  small-size real image dataset for which ground-truth motion is
provided. From this the ground-truth scales can be easily computed, and they range from $0.7$ to $43$ meters.
The relative rotations and translation directions are obtained following a standard approach based on the essential matrix factorization with a final bundle adjustment of camera pairs.

Table \ref{tab_real} shows the results obtained by all the variants of our method, namely FCB, MCB and N-MCB with threshold $\epsilon = 2^{\circ}$.
As in the case of simulated data, they all recover the translations norm accurately, and the best precision is achieved, on the average, by N-MCB.
\section{Conclusion} 

In this paper we have presented an in-depth study of the ESC problem, within the broader context of global structure from motion. 
After having provided
theoretical conditions under which such a problem has a unique solution, we have
presented an efficient algorithm to find it. 
The accuracy of our solution for computing the scaling
factors has been
demonstrated by means of synthetic and real experiments.

This method,  in combination with a motion synchronization technique
that works in SE(3) \cite{ArrigoniFR15},  constitutes the core of a global structure-from-motion pipeline that will be characterized experimentally in future work. 

On the theoretical side, we will explore the connection of our notion of ESC solvability and 
analogous concepts linked to the parallel (or bearing) rigidity \cite{Whi97,ZhaZel15}.
In this context we also aim at clarifying the separate role of $C$ and $B$ in Equation \eqref{eq_kr} in determining solvability.





{\small
\bibliographystyle{ieee}
\bibliography{references}

\begin{thebibliography}{10}\itemsep=-1pt

\bibitem{Arie12}
M.~Arie-Nachimson, S.~Z. Kovalsky, I.~Kemelmacher-Shlizerman, A.~Singer, and
  R.~Basri.
\newblock Global motion estimation from point matches.
\newblock In {\em International Conference on 3D Imaging, Modeling, Processing,
  Visualization and Transmission}, pages 81 -- 88, 2012.

\bibitem{ArrigoniFR15}
F.~Arrigoni, A.~Fusiello, and B.~Rossi.
\newblock Spectral motion synchronization in {SE(3)}.
\newblock {\em ArXiv e-prints}, 1506.08765, 2015.

\bibitem{isprsarchives-XL-5-63-2014}
F.~Arrigoni, B.~Rossi, F.~Malapelle, P.~Fragneto, and A.~Fusiello.
\newblock Robust global motion estimation with matrix completion.
\newblock {\em ISPRS - International Archives of the Photogrammetry, Remote
  Sensing and Spatial Information Sciences}, XL-5:63--70, 2014.

\bibitem{Brand04}
M.~Brand, M.~Antone, and S.~Teller.
\newblock Spectral solution of large-scale extrinsic camera calibration as a
  graph embedding problem.
\newblock In {\em Proceedings of the European Conference on Computer Vision},
  pages 262 -- 273, 2004.

\bibitem{disco}
D.~Crandall, A.~Owens, N.~Snavely, and D.~P. Huttenlocher.
\newblock Discrete-continuous optimization for large-scale structure from
  motion.
\newblock In {\em Proceedings of the IEEE Conference on Computer Vision and
  Pattern Recognition}, pages 3001 -- 3008, 2011.

\bibitem{EnqvistKO11}
O.~Enqvist, F.~Kahl, and C.~Olsson.
\newblock Non-sequential structure from motion.
\newblock In {\em Eleventh Workshop on Omnidirectional Vision, Camera Networks
  and Non-classical Camera}, pages 264 -- 271, 2011.

\bibitem{Govindu01}
V.~M. Govindu.
\newblock Combining two-view constraints for motion estimation.
\newblock In {\em Proceedings of the IEEE Conference on Computer Vision and
  Pattern Recognition}, pages 218 -- 225, 2001.

\bibitem{Govindu04}
V.~M. Govindu.
\newblock Lie-algebraic averaging for globally consistent motion estimation.
\newblock In {\em Proceedings of the IEEE Conference on Computer Vision and
  Pattern Recognition}, pages 684 -- 691, 2004.

\bibitem{horton}
J.~D. Horton.
\newblock A polynomial-time algorithm to find the shortest cycle basis of a
  graph.
\newblock {\em SIAM Journal on Computing}, 16(2):358 -- 366, 1987.

\bibitem{Jiang13}
N.~Jiang, Z.~Cui, and P.~Tan.
\newblock A global linear method for camera pose registration.
\newblock In {\em Proceedings of the International Conference on Computer
  Vision}, pages 481 -- 488, 2013.

\bibitem{Kahl08}
F.~Kahl and R.~Hartley.
\newblock Multiple-view geometry under the $l_{\infty}$-norm.
\newblock {\em IEEE Transactions on Pattern Analysis and Machine Intelligence},
  30(9):1603--1617, 2008.

\bibitem{bases}
T.~Kavitha, C.~Liebchen, K.~Mehlhorn, D.~Michail, R.~Rizzi, T.~Ueckerdt, and
  K.~Zweig.
\newblock Cycle bases in graphs: Characterization, algorithms, complexity, and
  applications.
\newblock {\em Computer Science Review}, 3(4):199 -- 243, 2009.

\bibitem{KatRao68}
C.~G. Khatri and C.~R. Rao.
\newblock Solutions to some functional equations and their applications to
  characterization of probability distributions.
\newblock {\em SankhyÄ: The Indian Journal of Statistics, Series A
  (1961-2002)}, 30(2):pp. 167--180, 1968.

\bibitem{Levi03}
N.~Levi and M.~Werman.
\newblock The viewing graph.
\newblock In {\em Proceedings of the IEEE Conference on Computer Vision and
  Pattern Recognition}, pages 518 -- 522, 2003.

\bibitem{MartinecP07}
D.~Martinec and T.~Pajdla.
\newblock Robust rotation and translation estimation in multiview
  reconstruction.
\newblock In {\em Proceedings of the IEEE Conference on Computer Vision and
  Pattern Recognition}, pages 1 -- 8, 2007.

\bibitem{Moulon13}
P.~Moulon, P.~Monasse, and R.~Marlet.
\newblock {Global Fusion of Relative Motions for Robust, Accurate and Scalable
  Structure from Motion}.
\newblock In {\em Proceedings of the International Conference on Computer
  Vision}, pages 3248 -- 3255, 2013.

\bibitem{OlssonE11a}
C.~Olsson and O.~Enqvist.
\newblock Stable structure from motion for unordered image collections.
\newblock In {\em Proc. of the Scandinavian conference on Image analysis},
  pages 524--535, 2011.

\bibitem{ozyesil2013stable}
O.~Ozyesil, A.~Singer, and R.~Basri.
\newblock Stable camera motion estimation using convex programming.
\newblock {\em SIAM Journal on Imaging Sciences}, 8(2):1120 -- 1262, 2015.

\bibitem{SinhaSS10}
S.~N. Sinha, D.~Steedly, and R.~Szeliski.
\newblock A multi-stage linear approach to structure from motion.
\newblock In {\em Proc. of the European Conference on Computer Vision}, pages
  267 -- 281, 2010.

\bibitem{real_data}
C.~Strecha, W.~von Hansen, L.~J.~V. Gool, P.~Fua, and U.~Thoennessen.
\newblock On benchmarking camera calibration and multi-view stereo for high
  resolution imagery.
\newblock In {\em Proceedings of the IEEE Conference on Computer Vision and
  Pattern Recognition}, pages 1 -- 8, 2008.

\bibitem{vismara97}
P.~Vismara.
\newblock Union of all the minimum cycle bases of a graph.
\newblock {\em Electronic Journal of Combinatorics}, 4(1), 1997.

\bibitem{Whi97}
W.~Whiteley.
\newblock Matroids from discrete geometry.
\newblock In J.~Bonin, J.~Oxley, and B.~Servatius, editors, {\em Matroid
  Theory}, AMS Contemporary Mathematics, pages 171--313. 1997.

\bibitem{Snavely14}
K.~Wilson and N.~Snavely.
\newblock Robust global translations with {1DSfM}.
\newblock In {\em Proceedings of the European Conference on Computer Vision},
  pages 61--75, 2014.

\bibitem{bayesian}
C.~Zach, M.~Klopschitz, and M.~Pollefeys.
\newblock Disambiguating visual relations using loop constraints.
\newblock In {\em Proceedings of the IEEE Conference on Computer Vision and
  Pattern Recognition}, pages 1426 -- 1433, 2010.

\bibitem{ZelFau96}
C.~Zeller and O.~Faugeras.
\newblock Camera self-calibration from video sequences: the {K}ruppa equations
  revisited.
\newblock Research Report 2793, INRIA, 1996.

\bibitem{ZhaZel15}
S.~{Zhao} and D.~{Zelazo}.
\newblock Bearing-only network localization: Localizability, sensitivity, and
  distributed protocols.
\newblock {\em ArXiv e-prints}, 1502.00154, 2015.

\end{thebibliography}
}

\end{document}